\documentclass[11pt]{article} 

\usepackage[T1]{fontenc}
\usepackage[utf8]{inputenc}

\usepackage{amsmath,amssymb,amsfonts}

\usepackage{graphicx}
\graphicspath{{./}{images/}} 
\usepackage{caption}
\usepackage{subcaption}
\usepackage{booktabs}
\usepackage{float}

\usepackage{algorithm}
\usepackage{algorithmicx}
\usepackage{algpseudocode}
\usepackage{listings}

\usepackage{xcolor}
\usepackage[title]{appendix} 
\usepackage[numbers,sort&compress]{natbib}
\usepackage{url}

\usepackage{hyperref}

\makeatletter
\AtEndDocument{\typeout{get arXiv to do 4 passes: Label(s) may have changed. Rerun}}
\makeatother

\title{Enhancing Automatic Modulation Recognition with a Reconstruction-Driven Vision Transformer under Limited Labels}

\author{
Hossein Ahmadi\\
The University of Akron\\
\texttt{ha168@uakron.edu}
\and
Banafsheh Saffari\\
The University of Akron\\
\texttt{bs273@uakron.edu}
\and
Sajjad Emdadi Mahdimahalleh\\
The University of Akron\\
\texttt{se67@uakron.edu}
\and
Mohammad Esmaeil Safari\\
Independent Researcher\\
\texttt{ms.safari@outlook.com}
\and
Aria Ahmadi\\
Independent Researcher\\
\texttt{ariaahmadi225@gmail.com}
}

\date{} 

\begin{document}
\maketitle

\begin{abstract}
Automatic modulation recognition (AMR) is critical for cognitive radio, spectrum monitoring, and secure wireless communication. 
However, existing solutions often rely on large labeled datasets or multi-stage training pipelines, which limit scalability and generalization in practical settings. 
In this paper, we propose a unified Vision Transformer (ViT) framework that integrates supervised, self-supervised, and reconstruction objectives. 
Our model combines a ViT encoder, a lightweight convolutional decoder, and a linear classifier, where reconstruction explicitly maps augmented signals back to their original forms, anchoring the encoder to fine-grained I/Q structures. 
This strategy encourages robust and discriminative feature learning during pretraining, while partial label supervision in fine-tuning ensures effective classification under data-scarce conditions. 
Extensive experiments on the RML2018.01A dataset show that our approach outperforms supervised CNN and ViT baselines in low-label regimes, approaches ResNet-level accuracy with only 15--20\% labeled data, and maintains strong performance across varying SNR levels. 
Overall, the proposed framework offers a simple yet powerful solution for efficient, generalizable, and label-efficient AMR.
\end{abstract}
\vspace{0.5em}

\section{Introduction}
The rapid expansion of wireless communication systems has heightened the need for reliable and efficient automatic modulation recognition (AMR) methods, particularly under limited supervision. Traditional deep learning models—most notably convolutional neural networks (CNNs)~\cite{xu2019cnn,zhang2019spwvd} and Vision Transformers (ViTs)~\cite{kong2023transformer}—have achieved impressive results on benchmark datasets such as RML2018~\cite{oshea2018over}. However, these fully supervised approaches rely heavily on large labeled datasets and often struggle with cross-domain generalization when faced with mismatched conditions such as hardware variability, channel fading, and signal-to-noise ratio (SNR) shifts~\cite{bu2020adversarial,wang2020zf}.

To reduce reliance on labeled data, recent studies have turned toward semi-supervised and self-supervised paradigms~\cite{liu2021selfcontrastive,wang2024sigmatch}. Teacher–student frameworks with exponential moving average updates~\cite{cai2025sscl,yin2024semi} have shown promise, yet they introduce architectural complexity and depend on carefully tuned augmentation schedules. Moreover, aggressive augmentations may distort subtle constellation patterns, which are critical for distinguishing high-order modulation schemes.

\textbf{In this paper, we propose a unified ViT-based AMR framework that combines unsupervised pretraining with semi-supervised fine-tuning.} The model consists of a Vision Transformer encoder, a lightweight convolutional decoder, and a linear classification head constructed by averaging patch embeddings. Unlike prior approaches, our framework emphasizes \emph{reconstructing augmented signals back to their original forms}. By explicitly forcing the model to recover the clean I/Q structure from noisy or warped augmentations, the encoder is encouraged to learn more accurate, fine-grained features that directly benefit downstream classification.

Our main contributions are summarized as follows:
\begin{itemize}
    \item We introduce a unified end-to-end ViT framework that jointly leverages supervised, self-supervised, and reconstruction objectives in a single-stage training pipeline, avoiding the complexity of teacher–student architectures.
    \item We propose a reconstruction-driven pretraining strategy where augmented signals are mapped back to their original forms, ensuring that learned representations remain faithful to signal-level structures while improving robustness to distortion.
    \item We demonstrate through extensive experiments that reconstruction-only pretraining achieves the strongest generalization under low-label regimes, outperforming contrastive-only and joint formulations. The proposed semi-supervised ViT surpasses CNN and supervised ViT baselines, and approaches ResNet-level accuracy with only 15--20\% of labels.
    \item We provide comprehensive evaluations across modulation classes and SNR levels, showing that the proposed framework remains resilient under noisy channel conditions and effectively captures high-order modulations. In addition, feature-space visualizations (e.g., t-SNE plots) confirm the emergence of well-separated clusters under low-label regimes, highlighting the discriminative power of the learned representations.
\end{itemize}

Overall, our framework offers a straightforward yet powerful approach to semi-supervised AMR by anchoring representation learning to the recovery of original signal structures. This novelty provides both improved classification accuracy and enhanced generalization, especially in data-scarce or noisy environments.

\section{Related Work}

Automatic Modulation Classification (AMC) has gained significant attention in recent years due to its critical role in military, cognitive radio, and non-cooperative communications. Traditional AMC methods relied on handcrafted features or likelihood-based estimations, which often failed in dynamic environments or under low signal-to-noise ratio (SNR) conditions. Recent advances have shifted focus toward deep learning and semi-supervised learning techniques to address challenges of limited labeled data and domain mismatch.

\subsection{Self-Supervised and Contrastive Learning}

Self-supervised contrastive frameworks such as SSCL-AMC \cite{cai2025sscl} and SemiAMC \cite{liu2021selfcontrastive} have shown great promise in extracting robust representations from unlabeled IQ signals. SSCL-AMC employs frequency-aware Transformer and LSTM layers to encode contrastive views generated via adversarial data augmentation. It then aggregates multiple classifiers using soft-voting for improved generalization. SemiAMC, inspired by SimCLR, uses rotation-based augmentation and a convolutional-LSTM encoder to learn discriminative embeddings, achieving high accuracy with minimal labeled data.

Wu et al. \cite{wu2024contrastive} extend this contrastive paradigm by integrating pseudo-labeling and stratified sampling into a two-stage framework. Their method fine-tunes the encoder on high-confidence pseudo-labels, improving label efficiency and balancing class distributions. Similarly, the RMAC framework \cite{liu2022augmented} combines confidence-based pseudo-label filtering and auxiliary classification to leverage both IQ and PA features. 

Zhao et al. \cite{zhao2024meta} introduced Meta Supervised Contrastive Learning (MSCL), which fuses meta-learning and supervised contrastive learning to adapt to new unseen modulation types with limited examples, maintaining inter-class separation and intra-class compactness.
Recent surveys have also emphasized the role of unsupervised methods in signal analysis. Ahmadi et al.~\cite{ahmadi2025unsupervised} provided a comprehensive review of Autoencoders and Vision Transformers for unsupervised time-series signal analysis across domains such as wireless communications, radar, biomedical engineering, and IoT, highlighting methodological advances, hybrid architectures, and open challenges in interpretability and domain generalization.

\subsection{Transformer-Based Architectures and Augmentation Strategies}

Kong et al. \cite{kong2023transformer} proposed a Transformer-based contrastive semi-supervised learning framework (TcssAMR), utilizing time-warping augmentation and hierarchical fine-tuning. A novel Convolutional Transformer Deep Neural Network (CTDNN) was designed to address position encoding and local dependency limitations of standard Transformers.

Shi et al. \cite{shi2024gafmae} introduced the GAF-MAE model, converting IQ signals into Gramian Angular Fields and using a masked autoencoder to extract meaningful representations through reconstruction. This vision-inspired approach works effectively with as little as 5\% labeled data.

\subsection{Semi-Supervised and Transfer Learning Frameworks}

Transfer learning has also been a popular technique to combat domain shift and data scarcity. TL-AMC \cite{wang2020zf} uses shared encoder weights between an autoencoder and classifier to extract transferable features across labeled and unlabeled domains. Bu et al. \cite{bu2020adversarial} introduced an adversarial transfer learning architecture (ATLA) that learns domain-invariant features using domain discriminators, achieving strong generalization even with 10\% labeled data. Similarly, Ahmadi et al.~\cite{ahmadi2025multiheart} proposed MultiHeart, a multimodal cardiac monitoring system that leverages a multi-task autoencoder framework to reconstruct missing modalities while performing classification. Although developed in the biomedical domain, MultiHeart demonstrates how multimodal fusion and transfer learning can enhance robustness and generalization, which is directly relevant to communication systems facing incomplete or noisy inputs.

In spectrum interference-based augmentation, Zheng et al. \cite{zheng2021spectrum} showed that injecting controlled frequency noise during STFT significantly improves AMC robustness. Similarly, AMRMIDAN \cite{deng2023modulation} used multimodal data and domain adversarial training to align representations and enhance classification under few-shot scenarios.

Wang et al. \cite{wang2024sigmatch} developed the Ensemble SigMatch (ESM) framework, combining multiview augmentation and consistency regularization, resulting in strong performance with limited labels.
Di et al.~\cite{di2024soamc} proposed SOAMC, a semi-supervised open-set recognition framework for automatic modulation classification (AMC). This method addresses the limitations of conventional AMC approaches in identifying unknown modulation types by combining self-supervised learning, consistency regularization, and open-set recognition techniques. SOAMC first pre-trains a feature extractor using a combination of labeled and unlabeled data with strong data augmentation strategies, such as rotation and flipping, to enhance intra-class compactness and inter-class separation. It then employs an open-set classifier that distinguishes known from unknown classes using an adaptive thresholding mechanism based on feature space analysis. Experimental results on benchmark datasets demonstrate that SOAMC achieves competitive accuracy while maintaining robustness in detecting unseen modulation types, showcasing its effectiveness in dynamic and non-cooperative wireless environments.

\subsection{Dataset Curation and Label Sanitization}

Beyond feature learning and model optimization, recent work has also addressed challenges related to data quality, label consistency, and device compliance. Mahjourian and Nguyen~\cite{mahjourian2025sanitizing} proposed the Vision-Language Sanitization and Refinement (VLSR) framework, which leverages CLIP-based vision-language embeddings and clustering to identify, correct, and consolidate noisy multi-label annotations in manufacturing datasets, thereby improving label consistency and overall dataset quality with minimal human intervention. Similarly, in the context of distributed medical imaging, Khankiki et al.~\cite{khankiki2025class} introduced a class imbalance-aware active learning framework using Vision Transformers within a federated learning setup for histopathological image classification. Their approach combines uncertainty-based sampling with data balancing strategies to enhance performance across non-iid client datasets while preserving data privacy. In another domain, aimed at improving reliability in wearable sensor studies,~\cite{aim2023compliance} developed a two-stage gradient-boosting classifier with post-processing to monitor compliance in real-time for the Automatic Ingestion Monitor v2 (AIM-2), achieving over 95\% accuracy and demonstrating its utility for food intake and physical activity research.

\subsection{Few-Shot and Open-Set Learning}

The emergence of new modulation types in the field has led to interest in fine-grained and open-set recognition. Feng et al. \cite{feng2025openset} proposed FOSSP, a framework using self-supervised pretraining with debiasing contrastive loss and synthetic unknown generation, capable of fine-grained open-set classification.

Zhou et al. \cite{zhou2024graph} introduced a graph-based semi-supervised approach for few-shot class-incremental AMC. It employs warm-up classification followed by label propagation over similarity graphs to enable incremental learning without catastrophic forgetting. In parallel, Ghajari et al.~\cite{ghajari2025hdc} presented an anomaly detection framework for IoT networks using hyperdimensional computing (HDC). Their approach, evaluated on the NSL-KDD dataset, achieved 91.55\% accuracy and effectively identified both known and unknown attack patterns, highlighting the broader applicability of open-set recognition strategies to communication security domains.

\subsection{Classical CNN and Modular Architectures}

Xu et al. \cite{xu2019cnn} explored various CNN structures on real signals and showed that transfer learning and denoising autoencoders improve performance under real-world noise. MsmcNet \cite{wang2022msmcnet} introduced a modular framework with dynamic feature processing modules based on task complexity and dataset size. This flexible architecture achieves strong performance in both AMC and transmitter identification tasks.

Zhang et al. \cite{zhang2022multiscale} used multi-scale convolutional networks to extract spatio-temporal features, combining center loss with cross-entropy to improve class separation. Kebaili et al. \cite{kebaili2024wdm} extended supervised classification to optical modulation types using quality metrics like Q-factor and BER.

Finally, Ali et al. \cite{ali2023highdim} proposed a supervised learning approach using PCA-extracted high-dimensional features, demonstrating robustness at low SNRs.

Recent AMC research trends emphasize hybrid frameworks that combine self-supervised representation learning, domain adaptation, and fine-grained open-set classification. Transformer encoders, contrastive losses, and multimodal fusion have become central themes in reducing reliance on labeled data while maintaining classification accuracy across dynamic and noisy communication environments.

\section{Signal Model and Problem Formulation}
In this section, the signal model adopted for automatic modulation classification (AMC) is introduced, followed by a formal definition of the learning objective.

\subsection{Signal Model}
In typical wireless communication systems, information is encoded into discrete complex symbols $s_k$ through modulation schemes. After digital-to-analog conversion and pulse shaping, the transmitted signal passes through a wireless channel and is corrupted by noise. The received baseband signal at the receiver can be represented as:
\begin{equation}
    r(n) = A(n) e^{j(\omega n + \theta)} x(n) + \sigma(n), \quad n = 0, 1, \dots, N-1
\end{equation}
where $r(n)$ is the received complex signal, $x(n)$ is the unknown modulated symbol, $A(n)$ denotes the channel gain, $\omega$ and $\theta$ represent the frequency and phase offsets respectively, and $\sigma(n)$ is additive white Gaussian noise (AWGN) with zero mean.

The complex-valued received signal is typically decomposed into its in-phase and quadrature components:
\begin{equation}
    I = \{ \Re[r(n)] \}_{n=0}^{N-1}, \quad Q = \{ \Im[r(n)] \}_{n=0}^{N-1}
\end{equation}
This I/Q representation forms the input to the neural network and captures the essential characteristics of most modulation schemes.

Alternatively, the received signal can be modeled as a hypothesis testing problem under $K$ candidate modulation classes. For the $k$-th class, the hypothesis is defined as:
\begin{equation}
    H_k: x_k(n) = s_k(n) + \omega_k(n), \quad n = 1, \dots, N
\end{equation}
where $x_k(n)$ is the received sample under hypothesis $H_k$, $s_k(n)$ is the transmitted signal, and $\omega_k(n)$ is AWGN.

\subsection{Problem Formulation}
Given a dataset of I/Q samples $\mathcal{D} = \{(x_i, y_i)\}$, the goal of AMC is to learn a classifier $g_\theta(\cdot)$ that maps input signals to modulation types $y \in \{1, 2, \dots, K\}$. The optimization objective is to maximize the posterior probability:
\begin{equation}
    \arg \max_{\theta} \, P\left( g_\theta(r_i) = y_k \mid y_\text{true} = y_k \right)
\end{equation}
where $\theta$ denotes the network parameters.

In semi-supervised AMC, the training dataset $\mathcal{D}$ is divided into a small labeled subset $\mathcal{D}_{\text{ld}} = \{(x_i, y_i)\}_{i=1}^N$ and a larger unlabeled subset $\mathcal{D}_{\text{ud}} = \{x_j\}_{j=1}^M$ where $M \gg N$. The loss function is composed of supervised and unsupervised components:
\begin{equation}
    \mathcal{L}_t = \alpha \sum_{x \in \mathcal{D}_{\text{ud}}} \mathcal{L}_u(x, \theta, \phi) + \beta \sum_{(x, y) \in \mathcal{D}_{\text{ld}}} \mathcal{L}_s(x, y, \theta, \phi)
\end{equation}
where $\mathcal{L}_s$ and $\mathcal{L}_u$ are supervised and unsupervised loss terms respectively, and $\phi$ represents the classifier head parameters. The weights $\alpha$ and $\beta$ balance the two terms.

This study considers 16 common modulation types and aims to classify each received sample into one of these classes using a hybrid semi-supervised learning strategy.

\section{Our Proposed Model}

\begin{figure}[htbp]
    \centering
    \includegraphics[width=\textwidth]{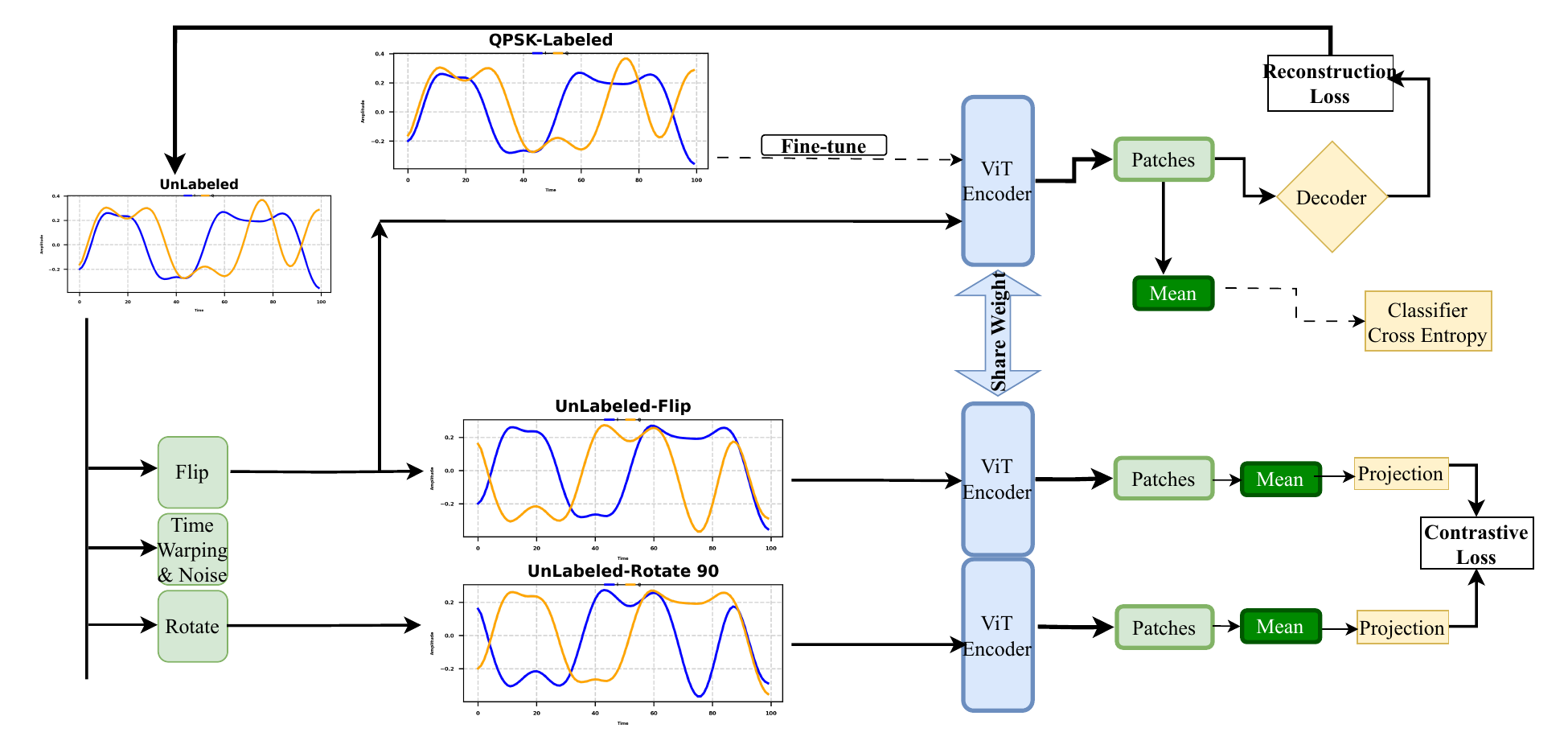}
    \caption{Overview of the main model architecture.}
    \label{fig:main_model}
\end{figure}

\subsection{Model Architecture and Training Strategy}

Our proposed model, illustrated in Figure~\ref{fig:main_model}, is composed of a Vision Transformer (ViT) encoder, a lightweight convolutional decoder, and a linear classification head. The design supports different pretraining scenarios by combining reconstruction, contrastive, and classification objectives in a flexible manner.

The reconstruction loss operates on the outputs of the decoder, encouraging the encoder to preserve signal-level features and retain the structure of the original I/Q samples. The contrastive loss is applied on projected latent representations of augmented signal pairs, enabling the encoder to learn invariant and discriminative features. Signal augmentations include Gaussian noise, scaling, rotation, horizontal and vertical flips, time warping, and magnitude warping, with only one transformation applied per sample at a probability of 1 to ensure diversity without excessive distortion.

During fine-tuning, the classification head is optimized using cross-entropy loss over varying fractions of labeled data (10\%, 15\%, and 20\%) spanning all 16 modulation classes. For unlabeled samples, pseudo-labels are generated from the model’s current predictions, and only those exceeding a confidence threshold of 0.8 are included in the classification loss with a reduced weighting factor (\texttt{pseudo\_label\_weight} = 0.5). This hybrid strategy allows the model to leverage both labeled and confidently inferred samples, enhancing discriminative learning without requiring additional manual annotations.

To balance contributions from each objective, scalar weights are applied to the reconstruction, contrastive, and classification losses. Patch embeddings are averaged instead of using a dedicated class token, ensuring that both the reconstruction and classification pathways remain anchored to the signal domain. Across all scenarios, this framework enables the encoder to learn semantically meaningful and robust representations while effectively utilizing both labeled and unlabeled data.

\subsection{Data Augmentation for Modulation Classification}

Deep learning–based automatic modulation classification (AMC) systems rely on large and diverse datasets to achieve robustness across varying channel and hardware conditions. In practice, however, collecting such extensive labeled datasets is often constrained by privacy concerns, annotation costs, and resource limitations. Data augmentation has therefore become a key strategy to enrich training data by generating transformed versions of existing signals while preserving their semantic meaning.

\begin{figure}[H]
    \centering
    \subfloat[Original Signal]{%
        \includegraphics[width=0.45\linewidth]{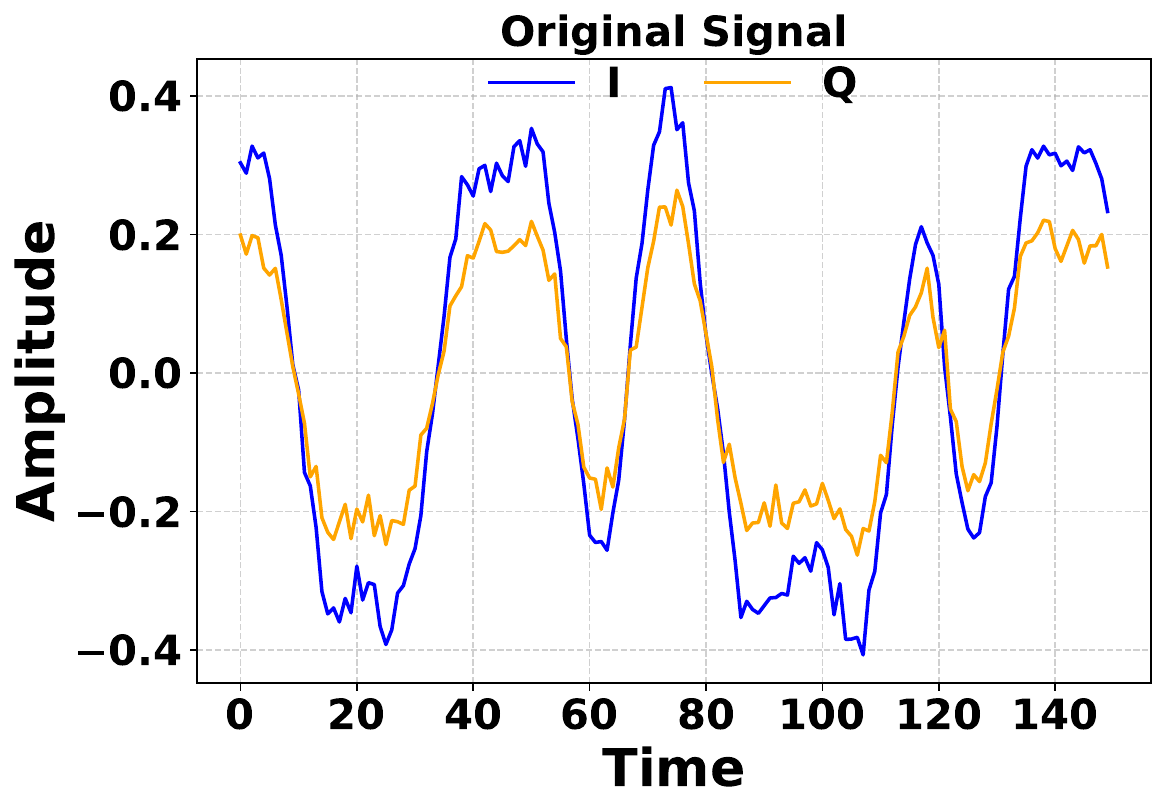}}
    \hfill
    \subfloat[Rotated Signal]{%
        \includegraphics[width=0.45\linewidth]{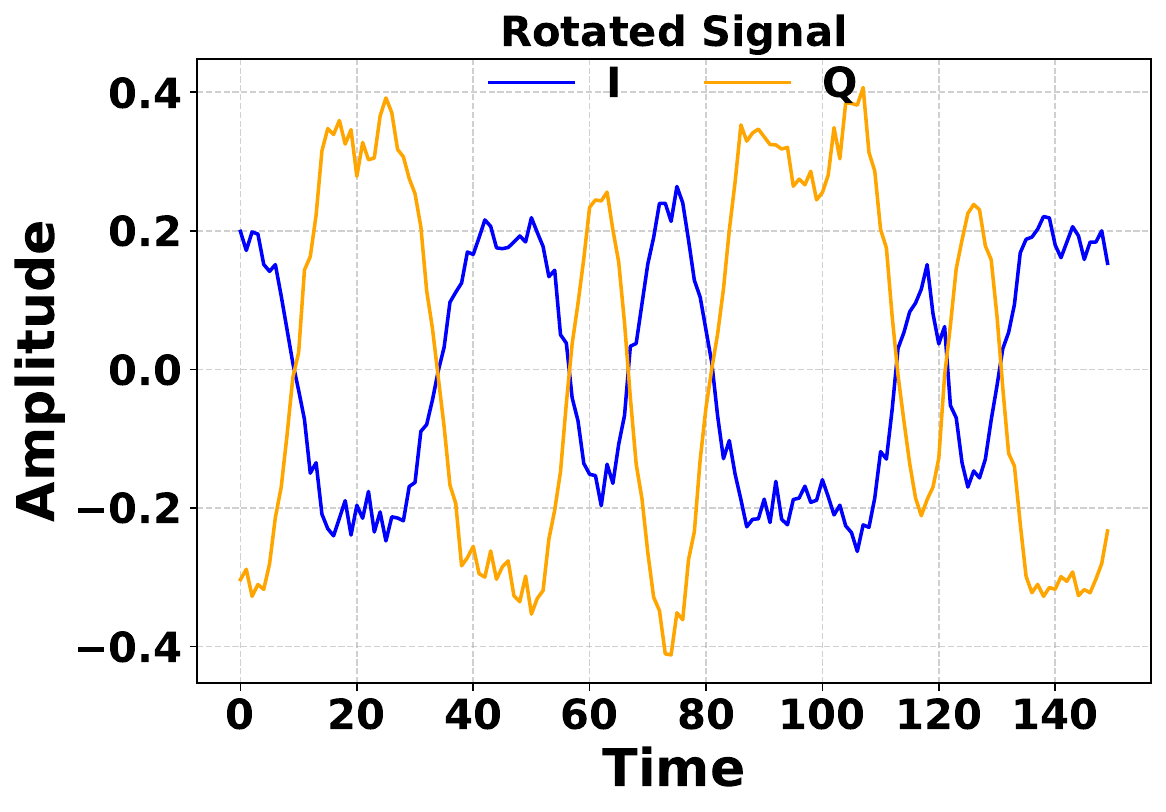}} \\[1ex]
    \subfloat[Flipped Signal]{%
        \includegraphics[width=0.45\linewidth]{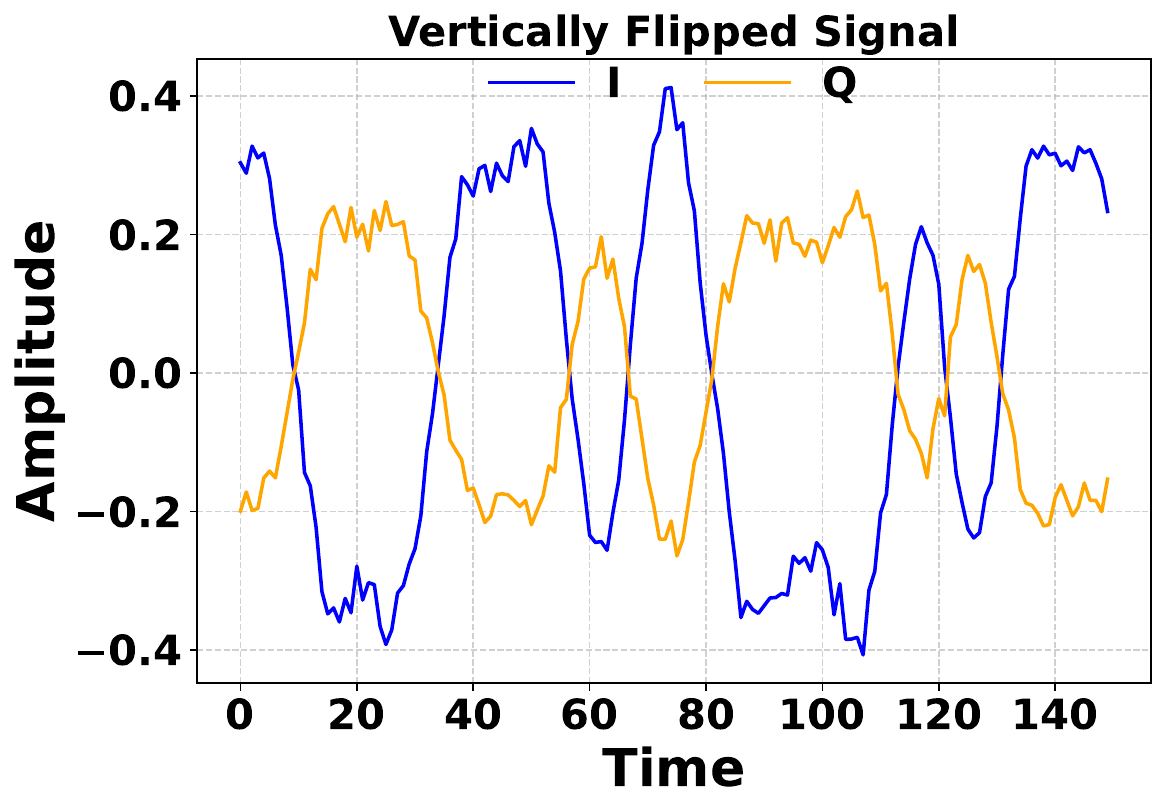}}
    \hfill
    \subfloat[Gaussian Noise]{%
        \includegraphics[width=0.45\linewidth]{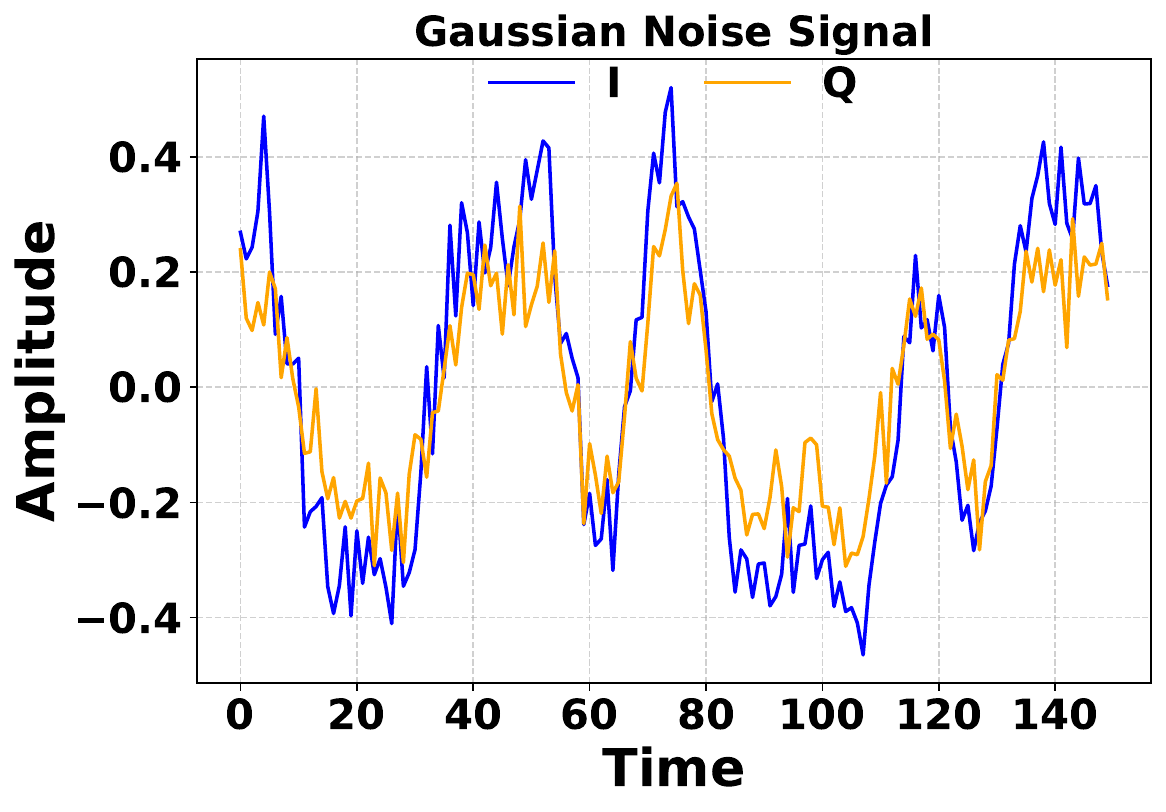}}
    \caption{Examples of original and augmented I/Q signals: (a) Original, (b) Rotated, (c) Flipped, and (d) Gaussian Noise.}
    \label{fig:signal_augmentations}
\end{figure}

Huang et al.~\cite{huang2020data} introduced several augmentation methods tailored for complex-valued I/Q signals. These include \textit{rotation}, in which each signal sample is rotated counterclockwise by fixed angles such as $0$, $\frac{\pi}{2}$, $\pi$, and $\frac{3\pi}{2}$, effectively shifting constellation points; \textit{flipping}, where horizontal or vertical reflections of the in-phase and quadrature components are applied; and \textit{additive Gaussian noise}, where controlled noise is injected to simulate realistic channel impairments. Such transformations introduce spatial diversity and improve generalization by exposing the model to a wider range of signal variations. Representative examples of these augmentations are shown in Figure~\ref{fig:signal_augmentations}.

Building on these classical methods, Pi et al.~\cite{pi2023improving} proposed augmentation techniques that specifically target temporal signal dynamics. \textit{Scaling} adjusts the amplitude of the signal using a Gaussian-distributed scalar multiplier, while \textit{magnitude warping} applies smooth nonlinear distortions generated through cubic spline interpolation. \textit{Time warping}, on the other hand, compresses or stretches the signal along the temporal axis via interpolated shifts. These methods expand variability in both amplitude and temporal resolution, while preserving the essential temporal and spectral structure of the original signal. Empirical studies demonstrated that such augmentations substantially improve classification performance, particularly under limited-data conditions, underscoring their importance for developing robust AMC systems.

\section{Experiments}

\subsection{Datasets}
We evaluate our proposed approach using the publicly available RML2018.01A dataset.

\textbf{RML2018.01A Dataset}~\cite{oshea2018over}: Provided by DeepSig Inc., this dataset consists of over 2.5 million complex I/Q samples, each represented as a 1024-length sequence, spanning 24 different modulation schemes over a wide range of SNRs (from $-20$\,dB to $+30$\,dB). It includes realistic wireless channel effects such as multipath fading, carrier frequency offsets, and additive white Gaussian noise (AWGN), offering a robust benchmark for evaluating modulation classification under diverse signal conditions.

In our experiments, we selected 16 modulation classes:
\begin{itemize}
    \item \texttt{BPSK}, \texttt{QPSK}, \texttt{8PSK}, \texttt{16APSK}, \texttt{32APSK}, \texttt{64APSK}, \texttt{128APSK},
    \item \texttt{16QAM}, \texttt{32QAM}, \texttt{64QAM}, \texttt{128QAM}, \texttt{256QAM},
    \item \texttt{AM-DSB-SC}, \texttt{AM-DSB-WC}, \texttt{FM}, and \texttt{GMSK}.
\end{itemize}

We restricted the SNR range to values between $-2$\,dB and $+21$\,dB, and sampled 1,000 examples per SNR per class. This setup ensures balanced data across modulation types and SNRs, facilitating consistent evaluation across supervised, self-supervised, and reconstruction-based training objectives.

\subsection{Experimental Setup}

The proposed framework is built on a Vision Transformer (ViT) encoder, paired with a lightweight decoder for reconstruction and a linear classification head. Patch embeddings are computed without using a class token, and their mean is used for both contrastive projection and classification.

\noindent\textbf{Dataset:}  
We evaluate our approach using the RML2018.01A dataset~\cite{oshea2018over}, filtered to 16 modulation classes and an SNR range of $-2$\,dB to $21$\,dB. For each class and SNR, 1,000 samples are selected, resulting in a total of 220,000 signals. The dataset is randomly shuffled and split into 70\% training, 10\% validation, and 20\% test sets. Each experiment is repeated 5 times with different shuffles, and results are reported as averages. Training is performed for 100 epochs in all scenarios.

\noindent\textbf{Training Scenarios:}  
We explore three different pretraining strategies, followed by fine-tuning:  
\begin{itemize}
    \item \textbf{Scenario 1 (Reconstruction Only)}: The model is pretrained using only the reconstruction loss.  
    \item \textbf{Scenario 2 (Reconstruction + Contrastive)}: Pretraining jointly optimizes reconstruction and contrastive losses.  
    \item \textbf{Scenario 3 (Contrastive Only)}: Pretraining is performed with contrastive loss alone.  
\end{itemize}
For fine-tuning, we experiment with different label ratios (10\%, 15\%, and 20\% of the training set). To establish an upper bound, we also evaluate classification performance when 100\% of the labels are available.  

\noindent\textbf{Data Augmentation:}  
To improve robustness, each training signal undergoes exactly one augmentation (selected from flipping, rotation, time-warping, scaling, magnitude warping, or Gaussian noise) with probability $1.0$, ensuring that every sample is augmented once per epoch.  

\vspace{0.5em}
\noindent\textbf{ViT Configuration:}
\begin{table}[H]
\centering
\begin{tabular}{|l|c|}
\hline
\textbf{Parameter} & \textbf{Value} \\
\hline
Input shape        & $(2, 512)$ \\
Patch size         & $(2, 16)$ \\
Embedding dimension & 64 \\
Number of layers   & 8 \\
Number of heads    & 8 \\
MLP dimension      & 64 \\
Projection output dim & 64 \\
Dropout            & 0.2 \\
Classifier token   & Not used (mean pooled patches) \\
Optimizer          & Adam \\
Learning rate      & 0.001 \\
Step size          & 15 \\
Learning rate decay factor & 0.90 \\
Weight decay       & 0.0 \\
\hline
\end{tabular}
\caption{Vision Transformer (ViT) encoder configuration and training hyperparameters.}
\end{table}

\vspace{0.5em}
\noindent\textbf{Baseline CNN Configuration:}
\begin{table}[H]
\centering
\begin{tabular}{|l|c|}
\hline
\textbf{Layer} & \textbf{Configuration} \\
\hline
Conv1 & 1$\times$7, 50 channels \\
Conv2 & 1$\times$7, 70 channels \\
Conv3 & 2$\times$7, 70 channels \\
Pooling & MaxPool2d, kernel 1$\times$2 \\
Dropout & 0.2 \\
FC1--FC4 & 4060$\rightarrow$512$\rightarrow$256$\rightarrow$80$\rightarrow$num\_classes \\
\hline
\end{tabular}
\caption{CNN baseline architecture for supervised classification.}
\end{table}

\vspace{0.5em}
\noindent\textbf{Baseline ResNet Configuration:}
\begin{table}[H]
\centering
\begin{tabular}{|l|c|}
\hline
\textbf{Component} & \textbf{Configuration} \\
\hline
Residual Block 1 & Input: 2 channels $\rightarrow$ 32 channels \\
Residual Stack   & 5 layers of 32 channels \\
Pooling          & MaxPool2d (2$\times$2), stride (1,2) \\
Fully Connected Layers & 256$\rightarrow$128$\rightarrow$128$\rightarrow$num\_classes \\
Dropout          & AlphaDropout (0.3) \\
Decoder (optional) & SimpleDecoder, input dim 128 \\
\hline
\end{tabular}
\caption{ResNet baseline architecture for supervised classification.}
\end{table}

\vspace{0.5em}
\noindent\textbf{Comparison Setup:}  
In addition to our ViT framework, we design two supervised baselines (CNN and ResNet) to provide a fair comparison. Thus, a total of three supervised models are evaluated under identical conditions. Results are reported across label ratios of 10\%, 15\%, 20\%, and full supervision (100\% labeled data).

\subsection{Results}

We evaluate the proposed framework under three distinct pretraining strategies, followed by fine-tuning with varying amounts of labeled data. All experiments are performed under identical conditions, with the input SNR range spanning from $-2$\,dB to $+20$\,dB to ensure a fair comparison across settings. The results demonstrate that pretraining with reconstruction alone (Scenario~1) consistently yields stronger performance compared to contrastive-only or joint reconstruction–contrastive setups. This highlights the importance of directly anchoring the encoder to signal-level structures through reconstruction. Furthermore, our ViT-based framework outperforms both CNN and ResNet baselines across all labeling ratios, confirming its effectiveness in low-label regimes.

\subsubsection{Scenario Comparison on RML2018.01A}
Table~\ref{tab:rml_scenarios} presents the classification accuracy achieved under the three pretraining scenarios using 10\%, 15\%, and 20\% labeled data. Scenario~1 (Reconstruction Only) consistently achieves the best results, reaching 68.21\% at 10\% labels and improving to 73.66\% with 20\% labels. Scenario~2 (Reconstruction + Contrastive) yields competitive performance but remains slightly below Scenario~1 across all labeling ratios, while Scenario~3 (Contrastive Only) provides the weakest performance despite modest gains with more labeled data. These results suggest that reconstruction-based pretraining provides more robust and generalizable feature representations, while also offering a more straightforward training pipeline than multi-objective formulations.

\begin{table}[H]
\centering
\begin{tabular}{|c|c|c|c|}
\hline
\textbf{Scenario} & \textbf{10\% Labels} & \textbf{15\% Labels} & \textbf{20\% Labels} \\
\hline
Scenario 1 (Reconstruction Only) & \textbf{68.21}$\pm$2\% & \textbf{71.01}$\pm$2\% & \textbf{73.66}$\pm$2\% \\
Scenario 2 (Reconstruction + Contrastive) & 66.40$\pm$2\% & 70.24$\pm$2\% & 71.41$\pm$2\% \\
Scenario 3 (Contrastive Only) & 53.41$\pm$2\% & 54.41$\pm$2\% & 56.41$\pm$2\% \\
\hline
\end{tabular}
\caption{Semi-supervised ViT performance under different pretraining objectives using 10\%, 15\%, and 20\% labeled data on the RML2018.01A dataset (SNR range $-2$\,dB to $+20$\,dB). Best values per column are highlighted in bold.}
\label{tab:rml_scenarios}
\end{table}

\subsubsection{Supervised and Semi-Supervised Model Comparison}
\begin{table}[H]
\centering
\begin{tabular}{|c|c|c|}
\hline
\textbf{Model} & \textbf{100\% Labels} & \textbf{Overall Accuracy (\%)} \\
\hline
CNN (Sup)     & Yes & 61.84 \\
ResNet (Sup)  & Yes & 78.50 \\
ViT (Sup)     & Yes & 68.21 \\
ViT (15\% Semi) & No  & 71.01 \\
\hline
\end{tabular}
\caption{Overall classification accuracy (\%) comparison of supervised CNN, ResNet, and ViT models trained with 100\% labels, and the proposed semi-supervised ViT with only 15\% labeled data.}
\label{tab:supervised_models_full}
\end{table}

\subsubsection{Per-Class Accuracy Comparison}
\begin{table}[H]
\centering
\resizebox{\textwidth}{!}{
\begin{tabular}{|c|c|c|c|c|}
\hline
\textbf{Modulation Class} & \textbf{CNN (Sup)} & \textbf{ResNet (Sup)} & \textbf{ViT (Sup)} & \textbf{ViT (15\% Semi)} \\
\hline
BPSK           & 99.84 & 99.96 & \textbf{100.00} & 99.91 \\
QPSK           & 90.50 & 91.55 & \textbf{97.31} & 87.88 \\
8PSK           & 93.44 & \textbf{97.48} & 95.73 & 96.38 \\
16APSK         & 80.78 & \textbf{94.85} & 89.12 & 85.85 \\
32APSK         & 76.16 & 80.44 & 84.25 & \textbf{82.78} \\
64APSK         & 31.87 & \textbf{65.59} & 54.87 & 64.32 \\
128APSK        & 28.94 & \textbf{70.71} & 21.07 & 58.10 \\
16QAM          & 46.36 & 79.98 & \textbf{82.12} & 76.52 \\
32QAM          & 24.08 & \textbf{77.80} & 59.39 & 67.58 \\
64QAM          & 23.83 & 41.20 & \textbf{44.91} & 24.64 \\
128QAM         & 13.89 & \textbf{57.55} & 16.57 & 48.07 \\
256QAM         & 23.84 & \textbf{45.12} & 21.50 & 40.20 \\
AM-DSB-SC      & \textbf{92.60} & 75.60 & 67.70 & 25.24 \\
AM-DSB-WC      & 62.48 & 78.12 & 59.90 & \textbf{78.51} \\
FM             & \textbf{100.00} & 99.92 & \textbf{100.00} & \textbf{100.00} \\
GMSK           & \textbf{99.96} & \textbf{99.96} & 99.42 & 99.85 \\
\hline
\end{tabular}
}
\caption{Per-class classification accuracy (\%) comparison across supervised CNN, ResNet, ViT, and semi-supervised ViT with 15\% labeled data. Best values are highlighted in bold.}
\label{tab:per_class_accuracy}
\end{table}

The per-class comparison in Table~\ref{tab:per_class_accuracy} highlights several trends. 
CNN performs reasonably well on simple modulations (BPSK, QPSK, 8PSK) but collapses on higher-order QAM/APSK. 
ResNet, with full supervision, consistently achieves the strongest performance overall, especially on challenging formats such as 16APSK (94.85\%), 128APSK (70.71\%), and 256QAM (45.12\%). 
The supervised ViT attains excellent results on low-order modulations such as QPSK (97.31\%) and 16QAM (82.12\%), but its performance sharply drops for higher-order QAM/APSK. 

The proposed semi-supervised ViT, despite using only 15\% labeled data, remains competitive in several categories. 
It surpasses ResNet in 32APSK (82.78\% vs.\ 80.44\%) and AM-DSB-WC (78.51\% vs.\ 78.12\%), and achieves comparable results on 64APSK (64.32\%) and 128QAM (48.07\%). 
Additionally, it matches supervised models with near-perfect recognition for FM and GMSK. 
However, it underperforms significantly in AM-DSB-SC and higher-order QAM (64QAM, 256QAM), reflecting the difficulty of learning fine-grained decision boundaries with fewer labels. 
Overall, these results demonstrate that the semi-supervised framework can approach, and in some cases exceed, fully supervised baselines, especially in classes where representation learning from unlabeled data is highly beneficial.
\subsubsection{Feature Representation Visualization}
\begin{figure}[H]
    \centering
    \includegraphics[width=0.8\linewidth]{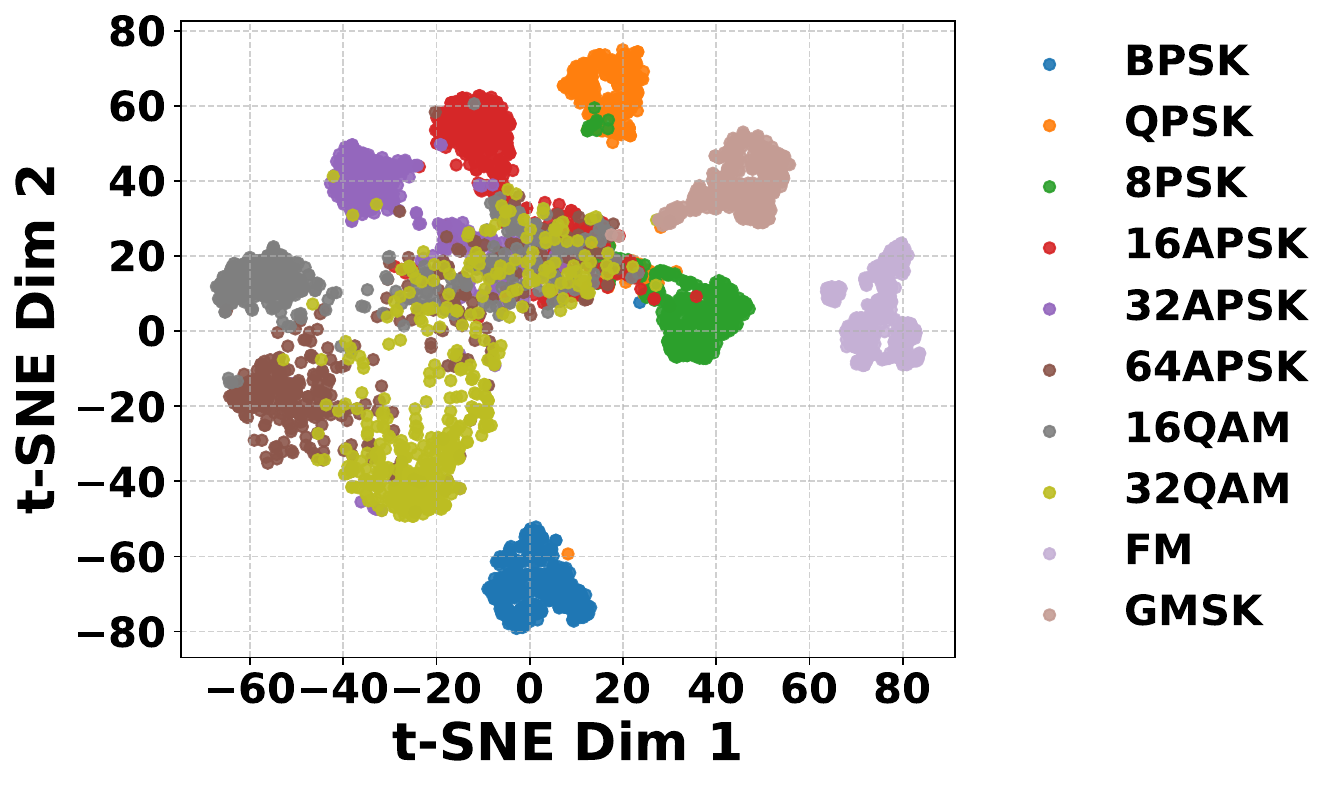}
    \caption{t-SNE visualization of Scenario~1 (reconstruction-only pretraining) for 10 modulation classes, demonstrating the formation of distinct feature clusters under low-label regimes.}
    \label{fig:tsne_progression}
\end{figure}

To further examine the quality of the learned representations, we visualize the latent space of the proposed framework using t-SNE. 
Figure~\ref{fig:tsne_progression} illustrates the embeddings obtained under Scenario~1 (reconstruction-only pretraining) with 15\% labeled data. 
The visualization shows that signals belonging to different modulation types form well-separated and compact clusters, 
highlighting the discriminative capability of the encoder even under low-label regimes.  

Notably, low-order modulations such as BPSK and QPSK exhibit tight and clearly separated clusters, 
while higher-order schemes (e.g., 32APSK and 64QAM) form more dispersed yet still distinguishable regions. 
This aligns with the per-class accuracy results reported in Table~\ref{tab:per_class_accuracy}, 
where higher-order constellations remain more challenging but still benefit from the reconstruction-driven pretraining. 
Overall, the t-SNE visualization provides intuitive evidence that anchoring the encoder to reconstruct augmented signals back to their original form 
leads to more structured and robust feature spaces that directly support downstream classification.

\subsubsection{Scenario-wise Per-Class Accuracy}
We further compare the three pretraining strategies under the setting of 15\% labeled data. 
For each scenario, we trained the model over five independent runs and report the average results on the held-out test set 
containing approximately 55{,}000 samples. Both the overall validation summaries and detailed per-class accuracies are shown below.

\begin{table}[H]
\centering
\resizebox{\textwidth}{!}{
\begin{tabular}{|c|c|c|c|}
\hline
\textbf{Modulation Class} & \textbf{Scenario 1 (Recon)} & \textbf{Scenario 2 (Recon + Contrastive)} & \textbf{Scenario 3 (Contrastive)} \\
\hline
BPSK           & 99.91 & 99.97 & 100.00 \\
QPSK           & 87.88 & 94.56 & 96.42  \\
8PSK           & 96.38 & 96.15 & 90.28  \\
16APSK         & 85.85 & 86.74 & 70.01  \\
32APSK         & 82.78 & 78.70 & 64.61  \\
64APSK         & 64.32 & 57.01 & 21.07  \\
128APSK        & 58.10 & 63.46 & 14.86  \\
16QAM          & 76.52 & 75.58 & 50.75  \\
32QAM          & 67.58 & 61.10 & 18.40  \\
64QAM          & 24.64 & 47.07 & 11.84  \\
128QAM         & 48.07 & 17.82 & 15.79  \\
256QAM         & 40.20 & 18.95 & 20.73  \\
AM-DSB-SC      & 25.24 & 57.42 & 14.59  \\
AM-DSB-WC      & 78.51 & 66.17 & 84.40  \\
FM             & 100.00 & 100.00 & 99.91 \\
GMSK           & 99.85 & 99.94 & 99.03 \\
\hline
\end{tabular}
}
\caption{Per-class classification accuracy (\%) comparison across the three pretraining scenarios with 15\% labeled data. 
Each scenario was trained five times, and results were evaluated on $\sim$55{,}000 test samples.}
\label{tab:scenario_per_class}
\end{table}

In terms of overall validation performance, Scenario~1 (Reconstruction-only) achieved an average accuracy of 71.01\% with a classification loss of 0.7383.  
Scenario~2 (Reconstruction + Contrastive) reached 70.24\% with a classification loss of 1.2866.  
Scenario~3 (Contrastive-only) dropped to 54.41\% with a classification loss of 2.0121.  
All three scenarios were evaluated on approximately 55{,}000 test samples.  

The per-class breakdown highlights several trends.  
Scenario~1 demonstrates balanced performance across modulation families, particularly maintaining relatively strong accuracies for high-order QAM and APSK classes 
(e.g., 128APSK at 58.10\% and 256QAM at 40.20\%).  
Scenario~2 benefits from the joint objectives, improving certain classes such as 64QAM (47.07\% vs.\ 24.64\% in Scenario~1) and AM-DSB-SC (57.42\% vs.\ 25.24\%).  
However, it also shows instability on very high-order modulations, with 128QAM and 256QAM dropping to 17.82\% and 18.95\%, respectively.  
Scenario~3, in contrast, performs competitively for simple constellations (BPSK, QPSK, FM, GMSK), yet its accuracy deteriorates sharply for higher-order schemes 
(32QAM at 18.40\%, 64APSK at 21.07\%, and 128APSK at only 14.86\%).  

Interestingly, AM-DSB-WC remains best captured by Scenario~3 (84.40\%), suggesting that contrastive-only training emphasizes 
distinct amplitude-modulated waveforms more effectively.  
Overall, the results indicate that reconstruction-based pretraining provides the most robust representations for complex modulations, 
while the joint approach can selectively enhance certain mid-complexity classes but struggles with consistency at the highest orders.

\subsubsection{SNR-wise Evaluation}
\begin{figure}[H]
    \centering
    \begin{subfigure}[b]{0.48\linewidth}
        \centering
        \includegraphics[width=\linewidth]{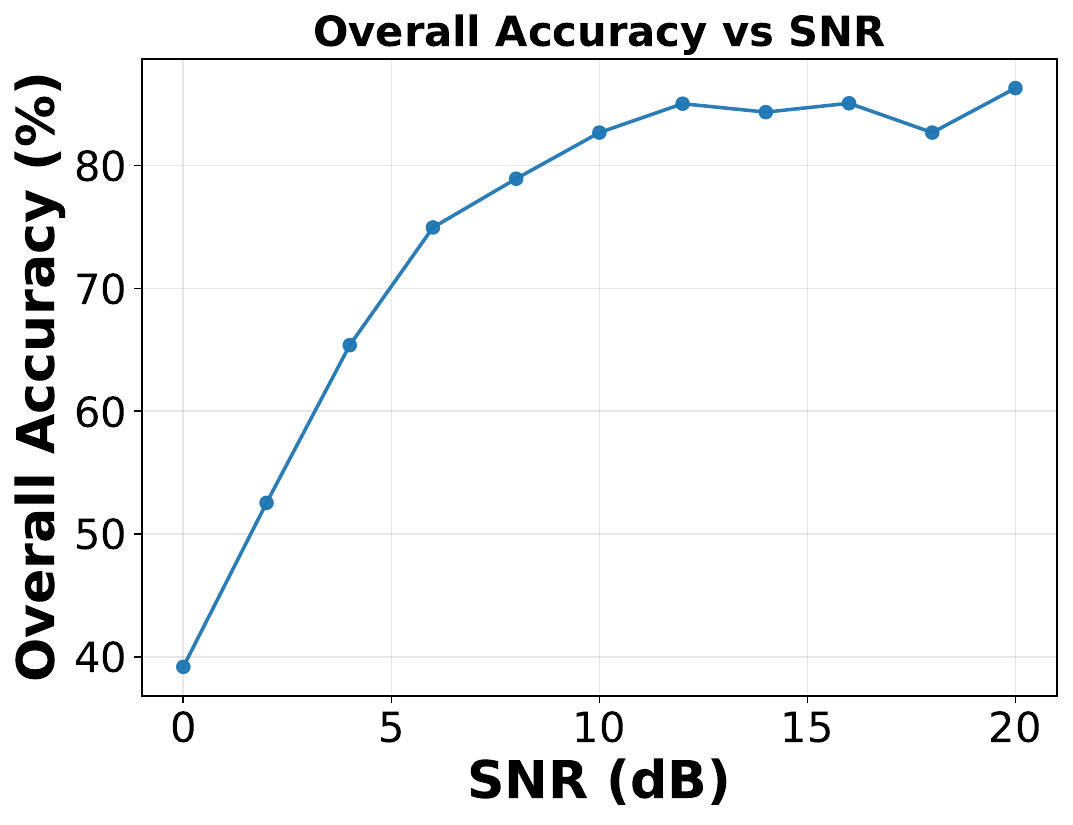}
        \caption{Overall classification accuracy across SNR levels.}
        \label{fig:snr_overall}
    \end{subfigure}
    \begin{subfigure}[b]{0.48\linewidth}
        \centering
        \includegraphics[width=\linewidth]{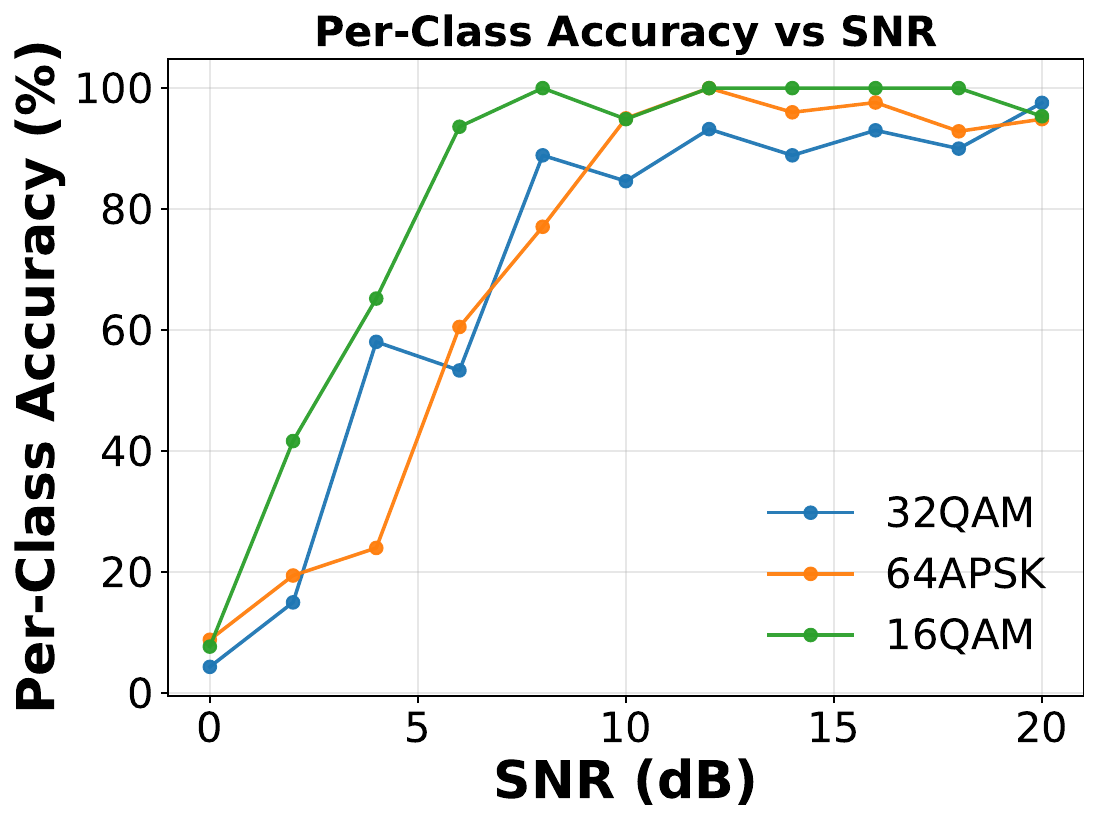}
        \caption{Per-class classification accuracy across SNR levels.}
        \label{fig:snr_per_class}
    \end{subfigure}
    \caption{Classification accuracy of the proposed model across different SNR levels ($-2$\,dB to $+22$\,dB) on the RML2018.01A dataset. 
    (a) Overall accuracy; (b) Per-class accuracy for three representative modulation classes. 
    Each SNR level was evaluated using 700 test samples.}
    \label{fig:snr_plots}
\end{figure}

We further evaluate the robustness of the proposed ViT framework across varying SNR levels. 
As shown in Figure~\ref{fig:snr_plots}, the model maintains strong performance even under low-SNR conditions, 
highlighting its resilience to channel noise and distortions. Notably, after an SNR value of 5\,dB, 
the model consistently achieves accuracy above 70\%, confirming its effectiveness in moderately noisy channels. 

In Figure~\ref{fig:snr_overall} (a), we present the overall classification accuracy across all modulation classes and SNR levels, 
while in Figure~\ref{fig:snr_per_class} (b), we show the per-class accuracy trends for three representative modulation types. 
This comparison illustrates how certain classes are more robust to noise than others. 
The evaluation was performed using 700 test samples at each SNR level to ensure reliability of the results. 

\section{Conclusion}
In this paper, we presented a unified Vision Transformer (ViT)--based framework for automatic modulation recognition (AMR) that jointly integrates supervised, self-supervised, and reconstruction objectives within a single training pipeline. Unlike existing approaches that rely on teacher–student architectures or multi-stage pretraining, our design emphasizes simplicity and efficiency by combining a ViT encoder with a lightweight decoder and linear classifier. This formulation allows the encoder to preserve fine-grained I/Q structures through reconstruction, while leveraging unlabeled data via contrastive learning and pseudo-labeling.

Extensive experiments on the RML2018.01A dataset demonstrate that reconstruction-only pretraining provides the most consistent and robust performance across scenarios, achieving competitive accuracy even when only 10--20\% of labels are available. The semi-supervised ViT framework not only surpasses CNN and ViT baselines under low-label regimes but also approaches fully supervised ResNet performance in several challenging modulation classes. Scenario-wise comparisons further reveal that reconstruction is particularly effective in stabilizing higher-order APSK and QAM recognition, while contrastive-only learning struggles to generalize beyond simpler constellations. Moreover, SNR-wise evaluations confirm that the proposed model maintains resilience under channel impairments, achieving accuracy above 70\% at moderate SNRs and demonstrating robustness to noise distortions.

Overall, the proposed framework advances the state of semi-supervised AMR by reducing dependence on large annotated datasets while maintaining scalability and interpretability. Future work will explore extensions to broader signal environments, including cross-dataset generalization, adversarial robustness, and real-time deployment on hardware platforms. Additionally, integrating domain-specific priors into the pretraining objectives may further enhance recognition of high-order modulations under low-SNR conditions.


\bibliography{Reference}

\begin{thebibliography}{10}

\bibitem{xu2019cnn}
Y.~Xu, D.~Li, Z.~Wang, Q.~Guo, and W.~Xiang, ``A deep learning method based on convolutional neural network for automatic modulation classification of wireless signals,'' {\em Wireless Networks}, vol.~25, no.~7, pp.~3735--3746, 2019.

\bibitem{zhang2019spwvd}
Z.~Zhang, C.~Wang, C.~Gan, S.~Sun, and M.~Wang, ``Automatic modulation classification using cnn with features fusion of spwvd and bjd,'' {\em IEEE Transactions on Signal and Information Processing over Networks}, vol.~5, no.~3, pp.~469--481, 2019.

\bibitem{kong2023transformer}
W.~Kong, X.~Jiao, Y.~Xu, B.~Zhang, and Q.~Yang, ``A transformer-based contrastive semi-supervised learning framework for automatic modulation recognition,'' {\em IEEE Transactions on Cognitive Communications and Networking}, vol.~9, no.~4, pp.~950--963, 2023.

\bibitem{oshea2018over}
T.~J. O’Shea, T.~Roy, and T.~C. Clancy, ``Over-the-air deep learning based radio signal classification,'' {\em IEEE Journal of Selected Topics in Signal Processing}, vol.~12, no.~1, pp.~168--179, 2018.

\bibitem{bu2020adversarial}
K.~Bu, Y.~He, X.~Jing, and J.~Han, ``Adversarial transfer learning for deep learning based automatic modulation classification,'' {\em IEEE Signal Processing Letters}, vol.~27, pp.~880--884, 2020.

\bibitem{wang2020zf}
Y.~Wang, G.~Gui, H.~Gacanin, T.~Ohtsuki, H.~Sari, and F.~Adachi, ``Transfer learning for semi-supervised amc in zf-mimo systems,'' {\em IEEE Journal on Emerging and Selected Topics in Circuits and Systems}, vol.~10, no.~2, pp.~231--243, 2020.

\bibitem{liu2021selfcontrastive}
D.~Liu, P.~Wang, T.~Wang, and T.~Abdelzaher, ``Self-contrastive learning based semi-supervised radio modulation classification,'' in {\em MILCOM 2021 - Special Session on Internet of Battlefield Things}, IEEE, 2021.

\bibitem{wang2024sigmatch}
H.~Wang, S.~Yang, Z.~Feng, and B.~Huang, ``Semi-supervised modulation classification via an ensemble sigmatch method,'' {\em IEEE Internet of Things Journal}, vol.~11, no.~20, pp.~32985--32997, 2024.

\bibitem{cai2025sscl}
Y.~Cai, D.~Li, S.~Wu, M.~Shao, S.~Hong, and H.~Sun, ``Sscl-amc: A self-supervised automatic modulation classification method via dynamic augmentation and ensemble learning,'' in {\em ICASSP 2025 - IEEE International Conference on Acoustics, Speech and Signal Processing}, IEEE, 2025.

\bibitem{yin2024semi}
L.~Yin, X.~Xiang, Y.~Liang, and K.~Liu, ``Modulation classification with data augmentation based on a semi-supervised generative model,'' {\em Wireless Networks}, vol.~30, pp.~5683--5696, 2024.

\bibitem{wu2024contrastive}
D.~Wu, J.~Shi, Z.~Li, M.~Du, F.~Liu, and F.~Zeng, ``Contrastive semi-supervised learning with pseudo-label for radar signal automatic modulation recognition,'' {\em IEEE Sensors Journal}, vol.~24, no.~19, pp.~30399--30410, 2024.

\bibitem{liu2022augmented}
H.~Liu and Z.~Zhu, ``Augmented semi-supervised learning for cnn based automatic modulation classification,'' in {\em IEEE ICCC}, IEEE, 2022.

\bibitem{zhao2024meta}
J.~Zhao, H.~Wang, S.~Peng, and Y.-D. Yao, ``Meta supervised contrastive learning for few-shot open-set modulation classification with signal constellation,'' {\em IEEE Communications Letters}, vol.~28, no.~4, pp.~837--841, 2024.

\bibitem{ahmadi2025unsupervised}
H.~Ahmadi, S.~E. Mahdimahalleh, A.~Farahat, and B.~Saffari, ``Unsupervised time-series signal analysis with autoencoders and vision transformers: A review of architectures and applications,'' {\em Journal of Intelligent Learning Systems and Applications}, vol.~17, no.~2, pp.~--, 2025.

\bibitem{shi2024gafmae}
Y.~Shi, H.~Xu, Y.~Zhang, Z.~Qi, and D.~Wang, ``Gaf-mae: A self-supervised automatic modulation classification method based on gramian angular field and masked autoencoder,'' {\em IEEE Transactions on Cognitive Communications and Networking}, vol.~10, no.~1, pp.~94--106, 2024.

\bibitem{ahmadi2025multiheart}
H.~Ahmadi, Y.~Zhang, and N.~H. Tran, ``Multiheart: Secure and robust heartbeat pattern recognition in multimodal cardiac monitoring system,'' {\em Electronics}, vol.~14, no.~15, p.~3149, 2025.

\bibitem{zheng2021spectrum}
Q.~Zheng, P.~Zhao, Y.~Li, H.~Wang, and Y.~Yang, ``Spectrum interference-based two-level data augmentation method in deep learning for automatic modulation classification,'' {\em Neural Computing and Applications}, vol.~33, no.~15, pp.~7723--7745, 2021.

\bibitem{deng2023modulation}
W.~Deng, X.~Wang, Z.~Huang, and Q.~Xu, ``Modulation classifier: A few-shot learning semi-supervised method based on multimodal information and domain adversarial network,'' {\em IEEE Communications Letters}, vol.~27, no.~2, pp.~576--580, 2023.

\bibitem{di2024soamc}
C.~Di, J.~Ji, C.~Sun, and L.~Liang, ``Soamc: A semi-supervised open-set recognition algorithm for automatic modulation classification,'' {\em Electronics}, vol.~13, no.~21, p.~4196, 2024.

\bibitem{mahjourian2025sanitizing}
N.~Mahjourian and V.~Nguyen, ``Sanitizing manufacturing dataset labels using vision-language models,'' {\em arXiv preprint arXiv:2506.23465}, 2025.

\bibitem{khankiki2025class}
M.~A.~L. Khankiki, M.~Mirzaeibonehkhatker, and S.~Esfandiari~Fard, ``Class imbalance-aware active learning with vision transformers in federated histopathological imaging,'' {\em Journal of Medicine and Medical Studies}, vol.~1, no.~2, pp.~65--73, 2025.
\newblock Published: 07 May, 2025.

\bibitem{aim2023compliance}
A.~names not provided~in abstract, ``Real-time wearable sensor compliance monitoring using a two-stage gradient-boosting classifier,'' in {\em 2023 International Conference on Electrical, Computer and Energy Technologies (ICECET)}, (Cape Town, South Africa), pp.~1--6, IEEE, Nov 2023.
\newblock Published: 22 January 2024.

\bibitem{feng2025openset}
Z.~Feng, H.~Pei, S.~Yang, and C.~Yang, ``Fine-grained open set signal modulation classification via self-supervised pre-training,'' {\em IEEE Transactions on Cognitive Communications and Networking}, 2025.

\bibitem{zhou2024graph}
X.~Zhou, P.~Qi, Q.~Liu, Y.~Ding, S.~Zheng, and Z.~Li, ``A graph-based semi-supervised approach for few-shot class-incremental modulation classification,'' {\em China Communications}, vol.~21, no.~11, pp.~88--103, 2024.

\bibitem{ghajari2025hdc}
G.~Ghajari, A.~Ghimire, E.~Ghajari, and F.~Amsaad, ``Network anomaly detection for iot using hyperdimensional computing on nsl-kdd,'' {\em arXiv preprint arXiv:2503.03031}, 2025.

\bibitem{wang2022msmcnet}
Y.~Wang, J.~Bai, Z.~Xiao, H.~Zhou, and L.~Jiao, ``Msmcnet: A modular few-shot learning framework for signal modulation classification,'' {\em IEEE Transactions on Signal Processing}, vol.~70, pp.~3789--3804, 2022.

\bibitem{zhang2022multiscale}
H.~Zhang, F.~Zhou, Q.~Wu, W.~Wu, and R.~Q. Hu, ``A novel automatic modulation classification scheme based on multi-scale networks,'' {\em IEEE Transactions on Cognitive Communications and Networking}, vol.~8, no.~1, pp.~97--111, 2022.

\bibitem{kebaili2024wdm}
R.~Kebaili, S.~Driz, and B.~Fassi, ``Supervised ml method based modulation formats classification for wdm systems,'' in {\em ISPA}, 2024.

\bibitem{ali2023highdim}
A.~K. Ali and E.~Ercelebi, ``Modulation format identification using supervised learning and high-dimensional features,'' {\em Arabian Journal for Science and Engineering}, vol.~48, no.~4, pp.~1461--1486, 2023.

\bibitem{huang2020data}
L.~Huang, W.~Pan, Y.~Zhang, L.~Qian, N.~Gao, and Y.~Wu, ``Data augmentation for deep learning-based radio modulation classification,'' {\em IEEE Access}, vol.~8, pp.~1498--1506, 2020.

\bibitem{pi2023improving}
S.~Pi, S.~Zhang, S.~Wang, B.~Guo, and W.~Yan, ``Improving modulation recognition using time series data augmentation via a spatiotemporal multi-channel framework,'' {\em Electronics}, vol.~12, no.~1, p.~96, 2023.

\end{thebibliography}


\end{document}